\documentclass[journal]{IEEEtran}

\usepackage{cite}
\ifCLASSINFOpdf
  \usepackage[pdftex]{graphicx}
\else
\fi
\usepackage{amsmath}
\usepackage{algorithmic}
\usepackage{array}
\usepackage{booktabs}
\usepackage{multirow}

\usepackage[colorlinks,linkcolor=red]{hyperref}

\begin{document}
\title{Unsupervised Joint Learning of Depth, Optical Flow, Ego-motion from Video}

\author{Jianfeng Li$^{3}$, Junqiao Zhao$^{*, 1, 2, 4}$, Shuangfu Song$^{3}$, Tiantian Feng$^{3}$
\thanks{*This work is supported by the National Key Research and Development Program of China (No. 2018YFB0105103, No. 2017YFA0603104), the National Natural Science Foundation of China (No. U1764261, No. 41801335, No. 41871370), the Natural Science Foundation of Shanghai (No. kz170020173571, No. 16DZ1100701) and the Fundamental Research Funds for the Central Universities (No. 22120180095).}
% <-this % stops a space
\thanks{$^{1}$The Key Laboratory of Embedded System and Service Computing, Ministry of Education, Tongji University, Shanghai}
\thanks{$^{2}$Department of Computer Science and Technology, School of Electronics and Information Engineering, Tongji University, Shanghai}%
\thanks{$^{3}$School of Surveying and Geoinfomatics, Tongji University, Shanghai}
\thanks{$^{4}$Insititute of Intelligent Vehicle, Tongji University, Shanghai}}

% make the title area
\maketitle

\begin{abstract}
Estimating geometric elements such as depth, camera motion, and optical flow from images is an important part of the robot's visual perception.
We use a joint self-supervised method to estimate the three geometric elements.
Depth network, optical flow network and camera motion network are independent of each other but are jointly optimized during training phase.
Compared with independent training, joint training can make full use of the geometric relationship between geometric elements and provide dynamic and static information of the scene.
In this paper, we improve the joint self-supervision method from three aspects: network structure, dynamic object segmentation, and geometric constraints.
In terms of network structure, we apply the attention mechanism to the camera motion network, which helps to take advantage of the similarity of camera movement between frames.
And according to attention mechanism in Transformer, we propose a plug-and-play convolutional attention module.
In terms of dynamic object, according to the different influences of dynamic objects in the optical flow self-supervised framework and the depth-pose self-supervised framework, we propose a threshold algorithm to detect dynamic regions, and mask that in the loss function respectively.
In terms of geometric constraints, we use traditional methods to estimate the fundamental matrix from the corresponding points to constrain the camera motion network.
We demonstrate the effectiveness of our method on the KITTI dataset.
Compared with other joint self-supervised methods, our method achieves state-of-the-art performance in the estimation of pose and optical flow, and the depth estimation has also achieved competitive results.
Code will be available at:\url{https://github.com/jianfenglihg/Unsupervised_geometry}.
\end{abstract}


\IEEEpeerreviewmaketitle

%%%%%%%%%%%%%%%%%%%%%%%%%%%%%%%%%%%%%%%%%%%%%%%%%%%%%%%%%%%%%%%%%%%%%%%%
\section{Introduction}
\IEEEPARstart{W}{ith} the rise of applications such as robots, autonomous driving, and augmented reality, the requirements for scene perception technology are also increasing. 
In various tasks of scene perception, the perception of geometric element information such as depth information, optical flow information, and camera motion information is particularly basic and important.
In addition, optical flow estimation and depth estimation are basic tasks in computer vision. 
Many computer vision research are based on accurate estimation of optical flow and accurate estimation of depth, such as action recognition\cite{sevilla2018integration}, video interpolation\cite{xu2019quadratic}, 3D target detection\cite{wang2019pseudo}, etc.

There are currently two types of methods for estimating depth, optical flow, and camera motion information from images. 
One is traditional methods based on nonlinear optimization, such as SLAM(simultaneous localization and mapping) and SFM(structure from motion), and the other is deep learning methods.
The disadvantage of traditional methods is that it is difficult to obtain dense estimation results and is easily affected by environmental factors such as illumination and texture. 
In addition, traditional methods will fail when motion is degenerate.
However, deep learning methods can quickly obtain dense estimation results in an end-to-end manner and are more robust.

Recent years, supervised deep learning techniques have been proposed to estimate depth, optical flow, and camera motion information.
The supervised training methods need to provide groud-truth, but the ground-truth of depth and optical flow are difficult to obtain.
Thus the limited training data restricts the generalization ability of the network.
More recently, self-supervised methods have been proposed which only uses videos tp complete the training, which greatly expands the types of training data.
The self-supervised methods realize image reconstruction by establishing the pixel correspondence between two frames, and convert the direct constraints on geometric elements in supervised learning into constraints on the accuracy of image reconstruction, so as to get rid of the dependence on the ground-truth.
Depth and pose establish pixel correspondence through the principle of multi-view geometry, and optical flow itself represents the pixel correspondence.

Research on supervised learning mainly focuses on improving the structure of convolutional neural networks, which greatly affects the estimation accuracy of depth, optical flow and camera motion.
And most self-supervised methods also directly inherit the network structure proposed by supervised methods.
In our work, we introduce the attention mechanism in NLP into the pose estimation network to improve the network's ability to capture the similarity of camera motion between multiple frames.

Optical flow itself can realize self-supervised training, and depth and pose can cooperate with each other through geometric relations to realize self-supervision. 
However, there is a more close geometric relationship between optical flow and depth, and camera movement. 
The joint self-supervised training that combines the three elements of depth, pose, and optical flow can make better use of the geometric constraints between them, such as epipolar constraints \cite{chen2019self}.
In our work, the eight-point method in the traditional method is used to establish geometric constraints between optical flow and pose.
In addition to geometric constraints, joint self-supervised methods also consider rigid/dynamic scene, occlusion regions and consistency between the depth and optical flow\cite{EPC++}, and we also proposed solutions to these problems.

The proposed framework is extensively evaluated on KITTI dataset\cite{KITTI}.
As elaborated in Sec.\ref{result}, our work outperforms other joint self-supervised learning methods.
In summary, the contributions of this paper lie in four aspects:
\begin{itemize}
  \item We introduce the attention mechanism into the pose estimation network, and propose a convolutional attention module, which explores the continuity and similarity of the motion between camera frames to improve the accuracy of pose estimation.
  \item We propose a new geometric constraint. By using the eight-point method in the epipolar geometry, the constraint between the optical flow and pose is established, which improve the accuracy of estimation.
  \item we propose a optical flow direction consistency constraint, which effectively improve the accuracy of optical flow prediction
  \item We propose a method for detecting moving objects, which effectively removes moving objects from the loss function.
\end{itemize}

%%%%%%%%%%%%%%%%%%%%%%%%%%%%%%%%%%%%%%%%%%%%%%%%%%%%%%%%%%%%%%%%%%%%%%%%%%%%%%%%%%%%%%%%%%%%%%%%%%%%%%%%%%%%%%%
\section{Related Works}
Estimating the depth of a single view and predicting 3D motion and optical flow based on images has long been a major problem in computer vision. 
Here, we summarize the most relevant works.

\subsection{Supervised learning of depth optical flow and ego-motion}
FlowNet\cite{dosovitskiy2015flownet} was the first to use convolutional neural networks to predict optical flow through a neural network structure of encoder and decoder.
And FlowNet produced a synthetic simulation data FlyingChair for training.
Inspired by the iterative optimization strategy in traditional methods, FlowNet2\cite{ilg2017flownet} stacks the encoder-decoder structure on the basis of FlowNet.
SpyNet\cite{spynet} is inspired by traditional methods from another aspect, constructing image pyramids during the training process, and estimating optical flow in a coarse-to-fine way.
Similar to SpyNet, PWC-Net\cite{sun2018pwc} also uses a pyramid strategy, but the difference is that PWC-Net constructs a feature pyramid instead of an image pyramid.
Compared with FlowNet2, PWC-Net has significantly improved the optical flow estimation accuracy and the amount of model parameters, and its parameter amount is 17 times smaller than that of FlowNet2. 
In our work, PWC-Net is also used as the backbone network of the optical flow prediction.

The first method of using deep learning technology to estimate pose is PoseNet\cite{kendall2015posenet}.
PoseNet returns the global pose, and then DeepVO\cite{wang2017deepvo} proposed a method to estimate the relative pose using a neural network.
DeepVO uses FlowNet network architecture to predict the relative pose of all adjacent frames, and then uses a recurrent neural network (RNN) to optimize the pose, which simulate the front and back ends of traditional SLAM.
Similar to DeepVo, parisotto\cite{parisotto2018global} also uses FlowNet as a network for preliminary prediction of relative pose. 
The difference is that parrisotto uses attention-based global optimization to simulate the back end of traditional SLAM.
However, the disadvantage of this method is that the relative pose cannot be estimated online, and the relative pose needs to be optimized after predicting the initial value of all frames.
We also applies the attention mechanism to relative pose regression, and our method is an online estimation method that can run in real time.

Regarding the supervised learning method of depth estimation, researchers are mainly devoted to the innovation of network structure.
Eigen\cite{eigen2014depth} firstly use neural networks to estimate depth, and proposed a network framework for estimating depth in a coarse-to-fine way.
DispNet\cite{mayer2016large} uses an encoder-decoder structure similar to FlowNet.
DeMoN\cite{ummenhofer2017demon} and Li et.al\cite{li2018undeepvo} predict camera motion and depth in a network at the same time, and combine these two tasks to achieve higher estimation accuracy.

\subsection{Unsupervised learning of Optical Flow}

\begin{figure*}[t]
  \centering
  \includegraphics[scale=0.6]{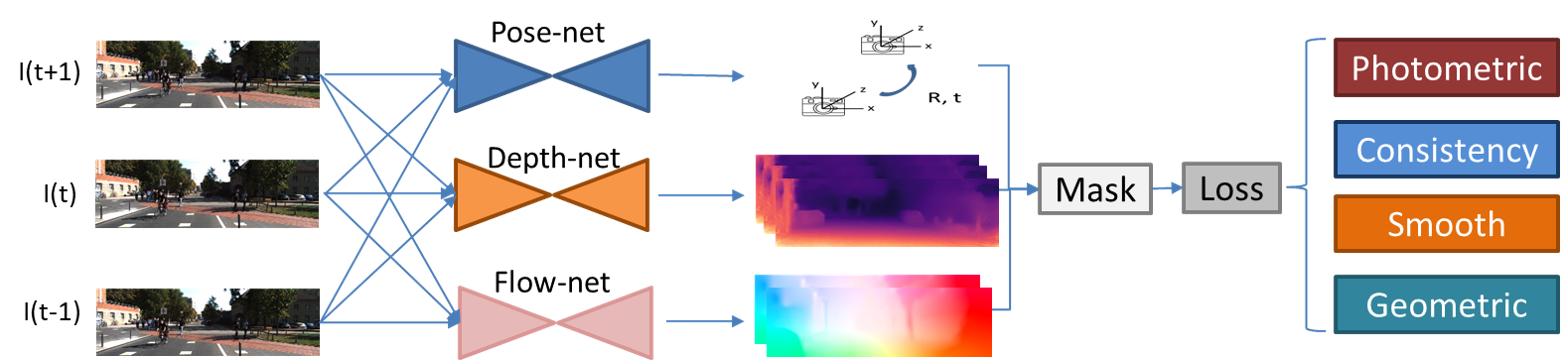}
  \caption{The pipeline of our framework. In the training phase, given three adjacent frames, the depth, optical flow, and pose networks respectively obtain the corresponding estimation results, and then various types of masks are calculated, and inappropriate areas are eliminated in the four types of loss functions.}
  \label{fig:overall}
\end{figure*}

Yu et al.\cite{jason2016back} and Ren et al.\cite{ren2017unsupervised} proposed self-supervised learning methods for optical flow. 
The optical flow and the next frame are used to reconstruct the previous frame. 
Through the constraints of image reconstruction and the global smoothness of optical flow, the self-supervised learning of optical flow is achieved.
In the smoothness constraint term, using the second-order regularization encourage the optical flow change in the neighborhood to tend to be linear, which can play a regularization role more efficiently\cite{meister2018unflow}\cite{janai2018unsupervised}\cite{wang2018occlusion}.
In addition to the improvement of the smooth regularization, another problem that needs to be solved is the occlusion problem.
Back2future\cite{janai2018unsupervised} added a branch to the optical flow estimation network to predict the occlusion area.
Unflow\cite{meister2018unflow} proposes a method to detect occlusion based on the consistency of the forward and reverse optical flow. 
Two adjacent frames of images are stacked in the positive and negative order and then input into the optical flow network to obtain the forward and reverse optical flow
OAFlow\cite{wang2018occlusion} also uses the forward and reverse optical flow to calculate the occlusion.
Different from UnFlow and OAFlow, Li et al.\cite{OAUnFlow} uses the forward and backward optical flow between three frames to calculate the occlusion, and achieves a higher optical flow estimation accuracy. 
We also adopted this method.

\subsection{Unsupervised learning of Depth and Ego-motion}
Clement Godard et al.\cite{godard2017unsupervised} and Garg et al.\cite{zhan2018unsupervised} used binocular images to achieve self-supervised learning of depth estimation.
There is an inverse relationship between the disparity and the depth in the binocular data.
Therefore, the corresponding relationship between the left and right image pixels can be obtained according to the depth, so that the corresponding relationship can be used to reconstruct the image to achieve self-supervision.
For monocular data, Zhou. et al\cite{zhou2017unsupervised} proposed a self-supervised framework for depth and camera motion estimation. 
In this framework, there are two networks that predict depth and camera motion respectively, and calculate pixel correspondences through camera motion and depth to achieve image reconstruction.
In order to improve the estimation accuracy of the depth and pose, Mahjourian\cite{mahjourian2018unsupervised} proposed the ICP (Iterative Nearest Point) geometric constraint. 
And bian et al\cite{bian2019unsupervised} proposed depth consistency constraint. 
In addition to geometric constraints, Vincent Casser et al\cite{casser2019unsupervised} used a pretrained semantic segmentation network to detect dynamic objects and predict the amount of motion, so as to make the geometric relationship of the dynamic region more rigorous.

\subsection{Unsupervised Joint Learning of Depth, Optical Flow and Ego-motion}
GeoNet\cite{yin2018geonet} unified the three geometric elements of depth, pose, and optical flow in a self-supervised learning framework.
GeoNet uses depth and pose to calculate the rigid optical flow, and then inputs the rigid optical flow into ResNet to estimate the residual optical flow. 
The final optical flow is the sum of the residual optical flow and the rigid optical flow.
DFNet\cite{zou2018df} uses three independent networks to predict depth, pose and optical flow respectively. 
In addition to the basic loss function, the consistency constraint of static optical flow and estimated optical flow is also added.
However, DFNet does not distinguish between dynamic and static regions when calculating optical flow consistency constraints. 
For this reason, Casser et al\cite{casser2019unsupervised} and Dai et al\cite{dai2020self} both use a pre-trained semantic segmentation network to obtain the mask of the dynamic region.
Although cc\cite{ranjan2019competitive} also uses neural networks to estimate dynamic and static regions, CC does not use a pre-trained network, but adds region segmentation to the self-supervised framework in a competitive and cooperative manner.
EPC++\cite{EPC++} does not use the network to estimate the dynamic area, but calculates it through rules that the scene flow estimated by camera motion and the scene flow estimated by optical flow are different in the dynamic area.
We also propose a rule-based occlusion detection algorithm.
In addition to the problem of dynamic objects, GLNet\cite{chen2019self} proposed to use epipolar geometric constraints in a self-supervised framework to improve the estimation accuracy. 
In our work, we use epipolar geometry in a new way.

%%%%%%%%%%%%%%%%%%%%%%%%%%%%%%%%%%%%%%%%%%%%%%%%%%%%%%%%%%%%%%%%%%%%%%%%%%%%%%%%%%%%%%%%%%%%%%%%%%%%%%%%%%%%%%%
\section{Method}
\subsection{Method overview}
The overall process of our method is shown in Fig.\ref{fig:overall}.
We solve the tasks of monocular depth prediction, optical flow and camera motion estimation by relying on three independent neural networks, referred to DepthNet, FlowNet and PoseNet respectively.
DepthNet estimates a pixelwise depth map from a single image, And the input of FlowNet and PoseNet are same, both are two images.
During the training phase, we utilize three adjacent frames $I_{t-1}$, $I_t$, $I_{t+1}$ in every epoch.
As Fig.\ref{fig:overall} demonstrates, we feed two pairs of images $I_{t}$, $I_{t-1}$ and $I_{t}$, $I_{t+1}$ into the FlowNet respectively by sharing weights, and then the bidirectional optical flow, the backward flow $F_b$ and the forward flow $F_f$ are obtained.
Here we define timestamp $t$ to $t-1$ as backward direction, and timestamp $t$ to $t+1$ as forward direction.
Similarly, we feed these image pairs into PoseNet and obtain the backward camera motion $R_b, t_b$ and forward motion $R_f, t_f$.
For depth, we estimate the depth map of each image, referred to as $D_{t-1}, D_t$ and $D_{t+1}$.
After that, we take in all the original images and network estimation results to calculate masks, and perform unsupervised training through loss functions.
There are four types of loss functions, namely photometric reprojection loss $L_{ph}$, smooth loss $L_s$, consistency loss $L_c$ and geometric loss $L_g$.
According to the adaptability of the loss function, we choose different masks to match and eliminate regions that do not fit the loss function.
The masks contains validity mask $M_v$, occlusion mask $M_o$ and dynamic object mask $M_d$.
The total loss function is as follows:
\begin{align}
  \label{total_loss}
  L &= \lambda_{ph}^{d}M_{v}M_{o}M_{d}L_{ph}^{d}+\lambda_{ph}^{f}M_{v}M_{o}L_{ph}^{f} \notag \\
    &+ \lambda_{c}^{d}M_{v}M_{o}M_{d}L_{c}^{d}+\lambda_{c}^{f}M_{d}L_{c}^{f}+\lambda_{c}^{df}M_{v}M_{o}M_{d}L_{c}^{df} \notag \\
    &+ \lambda_{s}^{d}L_{s}^{d}+\lambda_{s}^{f}L_{s}^{f} \notag \\
    &+ \lambda_{g}M_{v}M_{o}M_{d}L_{g}
\end{align}
Where $\lambda$ represents the weight of loss, the subscript $d$ and $f$ represent the loss related to depth and flow respectively.

\subsubsection{Geometric and appearance fundamental} \label{fundmental}
The key to achieving self-supervision for depth pose and optical flow are establishing pixel correspondence and image reconstruction.
Establish the pixel correspondence between image frames through the geometric elements estimated by the network, and then reconstruct the image by interpolation according to the pixel correspondence. 
The more accurate the image reconstruction, the more accurate the geometric elements estimation.

Given a pixel $p$ in the image $I_t$, consider camera intrinsics $K$ is known, the 3D scene point in camera coordinate system of timestamp $t$ can be back-projected as
\begin{equation} \label{back-projection}
  P_{t} = D_t(p_t)K^{-1}p_t
\end{equation}
According to the camera motion $R_f, t_f$, the coordinates of the 3D scene point can be transformed from the camera coordinate system at time $t$ to coordinates at time $t+1$.
And then the 3D scene point can be projected onto image $I_{t+1}$. 
Therefore, the process of establishing correspondence can be formulated as
\begin{equation} \label{reprojection}
 p_{t+1}^{dp}=K[R_f|t_f]D_t(p_t)K^{-1}p_t
\end{equation}
And projecting to image $t-1$ is the same process.
However, it is only suitable for static scenes that establishing point correspondence by coordinate transformation.
Correspondence of points suitable for dynamic objects can be established through optical flow, because the estimation of optical flow is based on appearance.
The calculation is as follows.
\begin{equation} \label{flow_corr}
 p_{t+1}^f=p_t+F_f(p_t)
\end{equation}
Therefore, when using the depth pose to calculate the pixel correspondence, the dynamic area should be eliminated.
We detect dynamic objects based on the difference between $p^{dp}_{t+1}$ and $p^f_{t+1}$ in the dynamic area.

After establishing the correspondence between pixels, the next step is according to the correspondence interpolating on $I_{t+1}$ and reconstructing $I_{t}$.
So we can get the image reconstruction error $E_{f}$ and $E_{b}$.
\begin{align} \label{eq:reconstrucgtion_error}
  E_{f}(p_t) &= |I_{t+1}(p_{t+1}^{f/dp})-I_{t}(p_t)| \notag \\
  E_{b}(p_t) &= |I_{t-1}(p_{t-1}^{f/dp})-I_{t}(p_t)| 
\end{align}

In this process, occlusion (pixels are not visible in another frame) will lead to incorrect interpolation results, which will mislead the optimization of the network during the training phase.
We detect the occlusion based on the difference of reconstruction error between the forward and backward direction, as presented in our previous work\cite{OAUnFlow}.
The methods of occlusion detection and dynamic object detection will be explained in \ref{OCC_DYN_mask}.

\subsection{Transformer Architecture in Pose Estimation} \label{PoseNet}
We add a convolutional attention module to the original pose estimation network based on the principle of the attention mechanism proposed by Transformer\cite{vaswani2017attention}.
The motivation here is to better exploit the advantages of the continuity and similarity of camera movement between frames.
Since the frame rate of the video is high enough and the camera motion is continuous, there is a high similarity between the forward camera motion $R_f$,$t_f$ and the backward camera motion $R_b$,$t_b$.
But it is difficult to describe this similarity relationship with a definite mathematical formula, so we choose to use neural network to explore the similarity of motion and increase the network's attention to similarity.
The advantage of the attention mechanism in Transformer is exploring the contextual relationship in natural language, so in principle it fits our needs that exploring the motion relationship between the previous and next frames.
The basic calculation formula of attention is:
\begin{equation}
  \label{attention}
  Attention(Q,K,V)=softmax(\frac{QK^{T}}{\sqrt{d_{k}}})V
\end{equation}
Where $Q$, $K$ and $V$ represent feature vectors. $Q$ and $K$ calculates the weight through the dot product operation, and weights V.

According to the principle of dot product attention, we designed the convolution attention module and inserted it into the original PoseNet \cite{zhou2017unsupervised}.
\begin{figure}[h]
  \centering
  \includegraphics[scale=0.4]{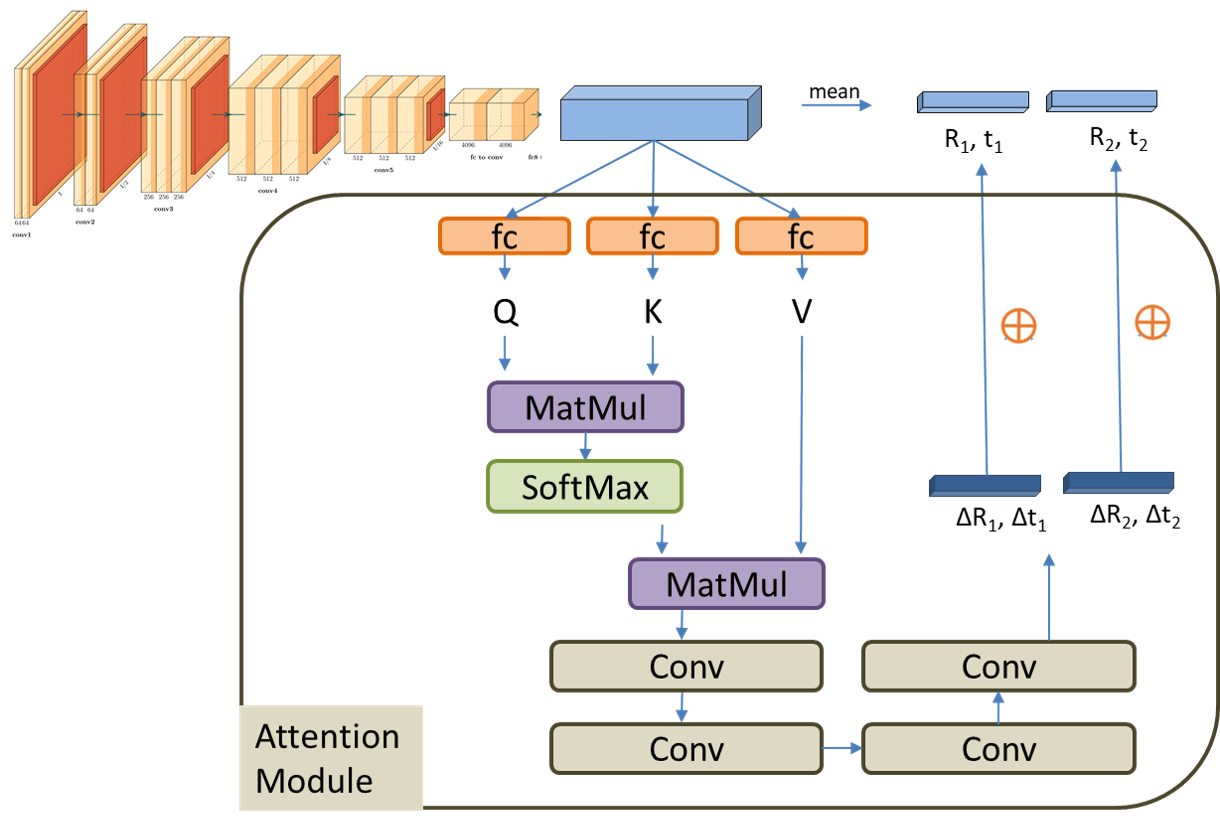}
  \caption{The architecture of PoseNet. The part in the box is the convolutional attention module proposed in this paper, and the upper part is the pose estimation network structure inherited from SFMLearner\cite{zhou2017unsupervised}.}
  \label{fig:atten_conv}
\end{figure}

As demonstrated in Fig.\ref{fig:atten_conv}, the feature map output by the last convolutional layer of the original network represents the description vectors of the forward and backward motion.
We use three fully-connected layers to obtain $Q$,$K$,$V$ features respectively from the motion description feature map, and then use the dot-product weighting method to obtain the fused feature map. 
After four layers of convolution, the correction amount of camera motion is obtained.
The final camera motion estimation value is the sum of the original estimation value and the correction amount.
In Sec.\ref{pose_result}, we demonstrate the improvement of pose estimation by the convolutional attention module.

\subsection{Occlusion and Dynamic Object Mask} \label{OCC_DYN_mask}
There are three types of masks in our work, namely validity mask, occlusion mask and dynamic object mask. 
Where validity mask can be computed analytically from depth and ego-motion estimates\cite{mahjourian2018unsupervised}.
Next, we mainly introduce occlusion masks and dynamic object masks.

\subsubsection{Occlusion mask}
In this paper, we use the occlusion detection algorithm in our previous work\cite{OAUnFlow}.
Because the forward and backward reconstruction errors are similar in the non-occluded area and have a large difference in the occluded area, we can detect the occluded area according to this prior rule.
As shown in the formula:
\begin{equation}
  M_{o} = 1-softmax(E_{b},E_{f}) > \delta  \ ? \ 0 : 1
\end{equation}
We refer readers to original paper for further details of the intuition and implementations.
In our experiment, we set $\delta$ to 0.48.

\subsubsection{Dynamic object mask}
As mentioned in Sec.\ref{fundmental}, the pixel correspondence can be obtained by coordinate transformation using the depth and pose, or can be obtained according to the optical flow.
However, for dynamic objects, using depth and pose to perform coordinate rigid-body transformation cannot get the correct pixel correspondence, but the optical flow method is still applicable.
Therefore, In our paper refers to the correspondence established by depth and pose as static optical flow $F_{rig}$, as shown in Eq.\ref{flow}.
\begin{align} \label{flow}
  &F_{rig} = p_{t+1}^{dp}-p_{t} \notag \\
  &F_{net} = p_{t+1}^{f}-p_{t}
\end{align}
Using the inconsistency between the static optical flow $F_{rig}$ and the estimated optical flow $F_{net}$ in the dynamic region, we can distinguish the dynamic region by using the threshold method as presented in \cite{meister2018unflow}.
\begin{equation}
  M_{d} = |F_{rig}-F_{net}|^{2} < \alpha_{1}(F_{rig}^2+F_{net}^2)+\alpha_{2} \ ? \ 1:0
\end{equation}
If the ratio of the optical flow difference to the optical flow is small enough, then it is considered to be a static area, otherwise it is considered to be a dynamic area.
In our experiments, we set $\alpha_{1}$ as 0.01 and $\alpha_{2}$ as 0.5.

\subsection{Optimization objectives}
In this section we describe the loss we use for the self-supervised learning, namely photometric constraints, smooth constraints, consistency constraints and geometric constraints.
These depth, optical flow and camera motion estimation tasks are linked together by these constraints.
\subsubsection{Photometric and smooth constraints}
Image reconstruction loss is widely used in self-supervised learning of optical flow and depth.
After using the estimated geometric entities to obtain the pixel correspondence between $I_{t}$ and $I_{t+1}$,  We can reconstruct $I_{t}^{'}$ from $I_{t+1}$.
In the same way, we can also reconstruct $I_{t}^{'}$ from $I_{t-1}$.
So as to get the reconstruction error $E_{f}$ and $E_{b}$, as Eq.\ref{eq:reconstrucgtion_error} shown.
If the reconstruction error is smaller, the geometric entity estimation is more accurate.
Therefore, the photometric loss is formulated as:
\begin{equation}
  L_{ph}=\beta_{1}|I_{t}^{'}-I_{t}|+\beta_{2}SSIM(I_{t}^{'},I_{t})
\end{equation}
where the L1 function is a kind of pixel-to-pixel loss, and SSIM \cite{wang2004image} can use neighborhood information, which helps the network to learn structural information.
In our experiment, we set the trade-off parameters $\beta_{1}=0.15$ and $\beta_{2}=0.85$.

Another loss term is the smooth regularization. The way of smooth regularization is similar to \cite{wang2018occlusion}, and we adopt second-order edge-aware formulation. 
Our smoothed loss function is defined as followings:
\begin{equation}
  L_{s}=\sum_{p}{|\nabla^{2}(G)|}e^{-\alpha|\nabla^{2}(I)|}
\end{equation}
where $G$ represents different geometry entities.

\subsubsection{Consistency constraints}
In our work, there are three consistency constraints, which are the consistency $L^{d}_{c}$ between the depth map, the consistency $L^{df}_{c}$ between the depth and the optical flow, and the consistency $L^{f}_{c}$ of the forward and backward optical flow.
The consistency between depth and flow is the same with \cite{zou2018df}, denoted as:
\begin{equation}
  \label{eq:dfc}
  L_{c}^{df}=\sum_{p_t}{|p_{t+1}^{f}-p_{t+1}^{dp}|}
\end{equation}
And the consistency of depth $L_{c}^{d}$ is the same with \cite{2019Unsupervised}, which is formulated as:
\begin{equation}
  \label{eq:d}
  L_{c}^{d}=\frac{|D_{t+1}(p_{t+1}^{dp})-D_{t+1}^{proj}(p_{t})|}{D_{t+1}(p_{t+1}^{dp})+D_{t+1}^{proj}(p_{t})}
\end{equation}
where $D_{t+1}^{proj}$ is the computed depth map of $I_{t+1}$ by transforming the point in camera coordinate $t$ into coordinate $t+1$.

In this paper, we propose a new optical flow consistency constraint based on the assumption that the directions of forward and backward optical flow are inverse to each other.
\cite{janai2018unsupervised} not only restricts the direction of forward and backward optical flow to be opposite to each other, but also requires the same magnitude, which is inaccurate.
Because even if the camera moves at a constant speed, due to the perspective projection, the forward and backward optical flow will not be equal.
But for static areas, the direction of optical flow is opposite.
Therefore, the consistency of optical flow is formulated as:
\begin{equation}
  \label{eq:fc}
  L_{c}^{f}=\sum_{p_t}{|\frac{F_{f}(p_{t})}{||F_{f}(p_t)||^{2}}+\frac{F_{b}(p_{t})}{||F_{b}(p_t)||^{2}}|}
\end{equation}

\subsubsection{Geometry constraint}
Epipolar geometry is the basic geometry of stereo vision and is widely used in classical computer vision method.
Epipolar geometry describes the relationship between corresponding epipolar points, that is, the corresponding epipolar point must lies on corresponding epipolar line on the other image.
The relationship writes
\begin{equation}
  \label{eq:epipolar}
  p'^{T}Fp=0
\end{equation}
where $p$ and $p'$ is corresponding points in two views respectively and F represents fundmental matrix which is composed of camera motion and intrinsics as follows.
\begin{equation}
  \label{eq:fundmental}
  F=K^{-T}[t]_{X}RK^{-1}
\end{equation}
GLNet \cite{chen2019self} directly substitute Eq.\ref{eq:fundmental} into Eq.\ref{eq:epipolar}, so that the optical flow and the pose are mutually constrained.
However, during experiment we found that the accuracy of optical flow estimation is difficult to further improve, indicating that the error of pose is more harmful to optical flow.
Therefore, this paper does not directly construct the epipolar geometric constraints like GLNet.
We use the eight-point algorithm \cite{longuet1981computer} to estimate the fundamental matrix $F_{est}$ from the optical flow in the manner of RANSAC.
And the corresponding points used for estimation are selected from static and non-occluded optical flow regions.
Subsequently, we use the pose to calculate the fundamental matrix $F_{cal}$ according to Eq.\ref{eq:fundmental}.
The resulting epipolar constraint loss writes
\begin{equation}
  \label{eq:geometric_loss}
  L_{g}^{epi}=|F_{est}-F_{cal}|
\end{equation}
Since the eight-point method and RANSAC are not diversified, the pose errors in this constraint will not be transmitted to the optical flow.

\section{Experiments}
In this section, we validate the effectiveness of our method on KITTI \cite{KITTI} dataset through extensive experiments of depth, optical flow and camera pose.
And we compare our method with existing joint self-supervised learning methods.

\subsection{Dataset for Training}
The KITTI dataset is captured in real world by a driving platform. 
We use KITTI raw data recordings to train our model, including 40864 samples in 65 training scenes and 3822 samples in 13 validation scenes.
The original images are down-sampled to 832x256 taking into account the GPU memory and batch size restriction. 
And then we evaluated optical flow on KITTI Flow 2012 and 2015 benchmarks, evaluated pose on KITTI odometry sequence 09 and 10.
And we evaluated the performance of depth on KITTI Eigen Split.

\subsection{Network architecture}
As Sec.\ref{PoseNet} described, PoseNet adds the attention module proposed in this paper on the basis of SfMLearner, and there are a total of 11 convolutional layers.
DepthNet is based on a encoder-decoder structure, and the encoder part is ResNet18 \cite{he2016deep}, the decoder part is the same with DispNet \cite{mayer2016large}.
The optical flow network is based on PWCNet \cite{sun2018pwc}.

\subsection{Training details}
We implement our end-to-end approach in PyTorch \cite{paszke2017automatic}. 
And our network is trained using Adam optimizer with $\beta_{1}$ = 0.9 and $\beta_{2}$ = 0.999.
The initial learning rate is set to be 1e-4, and is set to be 1e-5 after 200K iterations.
Taking into account the size of GPU memory, the batch size is set to be 8.
In this paper, we define $\lambda=[\lambda_{ph}^{d},\lambda_{ph}^{f},\lambda_{c}^{d},\lambda_{c}^{f},\lambda_{c}^{df},\lambda_{s}^{d},\lambda_{s}^{f},\lambda_{g}]$ as loss balancing hyper-parameters.

The whole training process consists of three stages.
In the first stage, the depth and pose self-supervised framework and the optical flow framework are trained independently.
So that each network has a better initial value.
The corresponding hyperparameter are $[1.0,1.0,0.1,0.01,0.0,10.0,0.5,0.0]$.
In the second stage, the two frames are jointly trained, and the dynamic object segmentation has a better initial result through the Cross-task consistency constraint.
The corresponding hyperparameter are $[1.0,1.0,0.1,0.01,1.0,10.0,0.5,0.0]$.
In the final stage, geometric constraints are added to the training framework, and The corresponding hyperparameter are $[1.0,1.0,0.1,0.01,1.0,10.0,0.5,0.1]$.
Beside, we calculate loss at 4 resolution-scales.

\subsection{results} \label{result}
\subsubsection{camera motion estimation} \label{pose_result}
We evaluate the performance of our method on the official KITTI visual odometry benchmark.
As a standard setting, we use the 09 and 10 sequence for testing, and we use the odometry metrics as in SfMLearner\cite{zhou2017unsupervised} and \cite{yin2018geonet}.
We compared with the traditional method, self-supervised method and joint self-supervised method.
Tab.\ref{table_Pose} demonstrates that our method achieves better odometry result than the traditional method or the self-supervised and joint self-supervised methods.

To better reveal the performance of the proposed method, we conducted ablation experiments.
Ours-base represents basic self-supervised method using photometric, smooth and depth-consistency constraints.
Ours-atten represents represents modifying the pose network structure according to the attention mechanism, yielding better result.
Ours-atten-joint means that the joint training of depth, pose and optical flow and the geometric constraints can effectively improve the accuracy of pose estimation.

\begin{table}[h]
  \renewcommand{\arraystretch}{1.3}
  \caption{Results of ego-motion estimation on the KITTI odometry seq.09 and seq.10}
  \label{table_Pose}
  \centering
  \begin{tabular}{ccc}
    \toprule
    Method & seq.09($mean \pm std$) & seq.10($mean \pm std$)\\
    \midrule
    ORB-Slam & $0.014 \pm 0.008$ & $0.012 \pm 0.011$ \\
    \hline
    DDVO & $0.045 \pm 0.108$ & $0.033 \pm 0.074$ \\
    Zhou & $0.016 \pm 0.009$ & $0.013 \pm 0.009$ \\
    Mahjourian & $0.013 \pm 0.010$ & $0.012 \pm 0.011$ \\
    Ranjan & $0.012 \pm 0.007$ & $0.012 \pm 0.008$ \\
    Monodepth2 & $0.017 \pm 0.008$ & $0.015 \pm 0.010$ \\
    \hline
    DF-Net & $0.017 \pm 0.007$ & $0.015 \pm 0.009$ \\
    GeoNet & $0.012 \pm 0.007$ & $0.012 \pm 0.009$ \\
    CC & $0.012 \pm 0.007$ & $0.012 \pm 0.008$ \\
    EPC++ M & $0.012 \pm 0.007$ & $0.013 \pm 0.008$ \\
    % EPC++ MS & $0.012 \pm 0.006$ & $0.012 \pm 0.008$ & 3\\
    GLNet & $0.011 \pm 0.006$ & $0.011 \pm 0.009$ \\
    \hline
    Ours-base & $0.013 \pm 0.008$ & $0.012 \pm 0.011$ \\
    Ours-atten & $0.011 \pm \mathbf{0.0058}$ & $0.011 \pm \mathbf{0.0073}$ \\
    Ours-atten-joint & $\mathbf{0.0098} \pm 0.0059$ & $\mathbf{0.0090} \pm 0.0074$ \\
    \bottomrule
    \end{tabular}
\end{table}

\subsubsection{optical flow estimation}

\begin{figure}[h]
  \centering
  \includegraphics[scale=0.7]{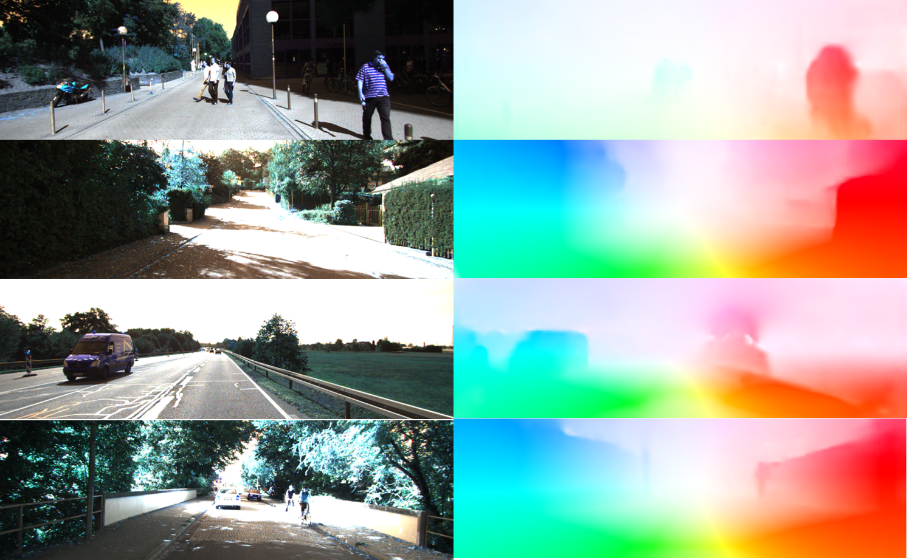}
  \caption{Qualitative results of optical flow estimation. left: input image, right: optical flow estimation results of our method.}
  \label{fig:flow}
\end{figure}
We evaluate the performance of our method on the Flow benchmark KITTI 2012\footnote{\url{http://www.cvlibs.net/datasets/kitti/eval_stereo_flow.php?benchmark=flow}} and KITTI 2015\footnote{\url{http://www.cvlibs.net/datasets/kitti/eval_scene_flow.php?benchmark=flow}}.
We report the performance using the average end-point error (EPE) over non-occluded regions (noc), occluded regions(occ) and overall regions (all).
We train the network on the KITTI RAW dataset, and evaluated on the training and testing set of KITTI 2012 and 2015.
The detailed evaluation results of our method can be obtained from the official website.

Qualitative results are provided in Fig.\ref{fig:flow} and the quantitative results are shown in Tab.\ref{table_flow}.
We compare with several self-supervised methods and joint self-supervised SOTA methods.
From the table we can see that our method is comparable to other unsupervised learning methods with slight improvement but still behind the state-of-the-art supervised methods Selflow, which adopted a different idea to deal with occlusion, instead of excluding occlusion but training the network to guess the optical flow of occlusion.
Compared with joint self-supervised methods, we achieved SOTA result.
And espetially, comparing with GLNet, we use epipolar geometry in different ways to reduce the influence of pose error on optical flow, yielding better result.

\begin{table*}[h]
  \renewcommand{\arraystretch}{1.3}
  \caption{Results of Optical Flow estimation on KITTI 2015 and KITTI 2012}
  \label{table_flow}
  \centering
  \begin{tabular}{cc|c|ccc|ccc}
    \toprule
    \multirow{3}{*}{Method} & \multicolumn{2}{c}{KITTI 2012}  & \multicolumn{6}{c}{KITTI 2015}\\ 
    \cmidrule(r){2-3}
    \cmidrule(r){4-9}
    & train & test & \multicolumn{3}{c}{train} & \multicolumn{3}{c}{test} \\
    \cmidrule(r){2-2}
    \cmidrule(r){3-3}
    \cmidrule(r){4-6}
    \cmidrule(r){7-9}
    & all & all & noc & occ & all & bg & fg & all \\
    \hline
    DSTFlow-ft & 10.43 & 12.40 & 6.96 & - & 16.79 & - & - & 39.00\% \\
    OAFlow-ft & 3.55 & 4.20 & - & - & 8.88 & - & - & 31.20\% \\
    UnFlow & 3.29 & - & - & - & 8.10 & - & - & - \\
    Back2Future-ft & - & - & 3.22 & 19.11 & 6.59 & - & - & 22.94\% \\
    DDFlow-ft & 2.35 & 3.0 & 2.73 & 24.68 & 5.72 & - & - & 14.29\%\\
    Selflow & \textbf{1.69} & \textbf{2.2} & \textbf{2.40} & \textbf{19.68} & \textbf{4.84} & - & - & \textbf{14.19\%} \\
    \hline
    GeoNet & - & - & - & - & 10.81 & - & - & - \\
    DFNet & 3.54 & 4.4 & - & - & 8.98 & - & - & 25.70\% \\
    CC-ft & - & - & - & - & 5.66 & - & - & 25.27\% \\
    EPC++ & 2.3 & 2.6 & 3.84 & \textbf{15.72} & 5.84 & 20.61\% & 26.32\% & 21.56\% \\
    GLNet & - &  - & 4.86 & - & 8.35 & - & - & - \\
    Ours & \textbf{1.97} & \textbf{2.2} & \textbf{3.03} & 17.52 & \textbf{5.66} & \textbf{18.57\%} & \textbf{24.02\%} & \textbf{19.48\%} \\
    \bottomrule
    \end{tabular}
\end{table*}

\subsubsection{monocular depth estimation}
\begin{figure}[h]
  \centering
  \includegraphics[scale=0.74]{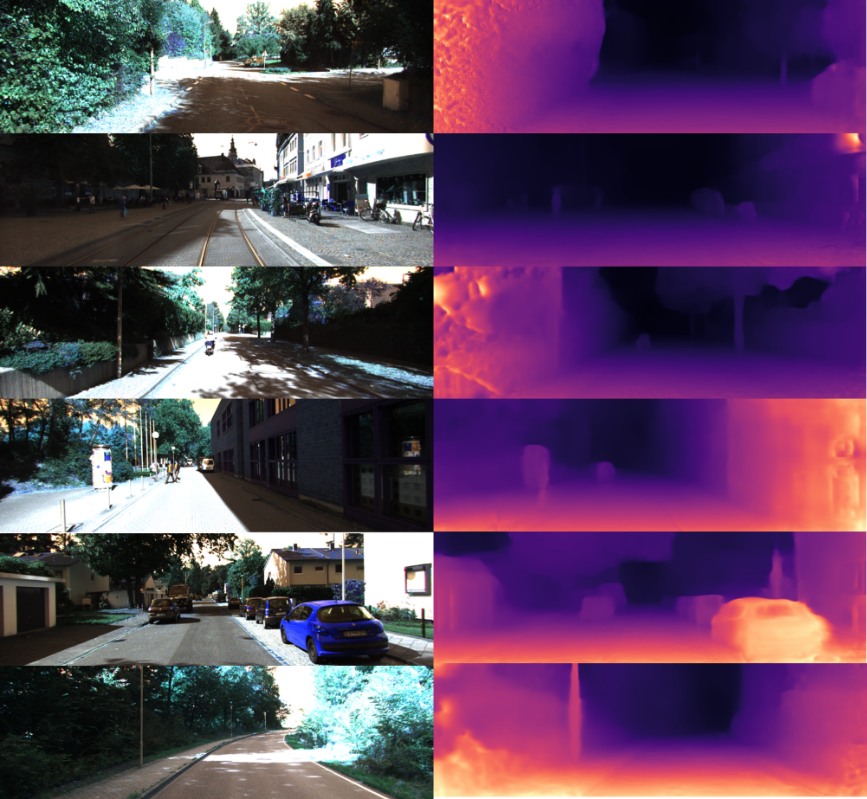}
  \caption{Qualitative results of depth estimation. left: input image, right: depth estimation results of our method.}
  \label{fig:depth}
\end{figure}

We also evaluate the performance of our method using the Eigen split of the raw KITTI dataset.
We compare the performance of the proposed framework with self-supervised methods and joint self-supervised methods.
Tab.\ref{table_depth} illustrate that our method is competitive to other self-supervised methods and joint self-supervised methods, but still behind the state-of-the-art methods.

In order to better reveal the performance of the proposed method, we conducted a series of ablation experiments.
As Tab.\ref{table_depth} shown, ours-base and ours-atten are same with methods described in Sec.\ref{pose_result}.
And -dfc means adding Cross-task consistency constraint $L_{c}^{df}$ described in Eq.\ref{eq:dfc}, after using the consistency loss the performance is improved.
-g means adding epipolar geometric constraint $L_{g}$, after that the performance is further improved.

\begin{table*}[h]
  \renewcommand{\arraystretch}{1.3}
  \caption{Results of depth estimation on the KITTI Eigen Split}
  \label{table_depth}
  \centering
  \begin{tabular}{cccccccc}
    \toprule
    \multirow{2}{*}{Method} & \multicolumn{4}{c}{Lower the better} & \multicolumn{3}{c}{Higher the better} \\
    \cmidrule(r){2-5}
    \cmidrule(r){6-8}
    & Abs Rel & Sq Rel & RMSE & RMSE log & $\delta<1.25$ & $\delta<1.25^2$ & $\delta<1.25^3$ \\
    \midrule
    SC-SfMLearner & 0.137 & 1.089 & 5.439 & 0.217 & 0.830 & 0.942 & 0.975 \\
    Monodepth2 & 0.132 & 1.044 & 5.142 & 0.210 & 0.845 & 0.948 & 0.977 \\
    \hline
    GeoNet-ResNet(update) & 0.149 & 1.060 & 5.567 & 0.226 & 0.796 & 0.935 & 0.975 \\
    % DFNet(cs+k)  & 0.146 & 1.182 & 5.215 & 0.213 & 0.818 & 0.943 & 0.978 \\
    DFNet(k) & 0.150 & 1.124 & 5.507 & 0.223 & 0.806 & 0.933 & 0.973 \\
    CC(k) & 0.140 & 1.070 & 5.326 & 0.217 & 0.826 & 0.941 & 0.975 \\
    % CC(cs+k) & 0.139 & 1.032 & 5.199 & 0.213 & 0.827 & 0.943 & 0.977 \\
    EPC++ & 0.141 & 1.029 & 5.350 & 0.216 & 0.816 & 0.941 & 0.976 \\
    GLNet & \textbf{0.135} & 1.070 & \textbf{5.230} & \textbf{0.210} & \textbf{0.841} & \textbf{0.948} & \textbf{0.980} \\
    \hline
    Ours-base & 0.147 & 1.0638 & 5.594 & 0.228 & 0.802 & 0.934 & 0.974 \\
    Ours-atten & 0.144 & 0.984 & 5.503 & \textbf{0.225} & 0.808 & 0.938 & \textbf{0.975} \\
    ours-atten-dfc & 0.140 & 0.971 & \textbf{5.375} & 0.228 & 0.817 & \textbf{0.939} & 0.973 \\
    Ours-atten-dfc-g & \textbf{0.138} & \textbf{0.970} & 5.460 & 0.231 & \textbf{0.820} & 0.938 & 0.971 \\
    % Ours-atten-dfc-8p-tri & 0.140 & 1.001 & 5.591 & 0.230 & 0.816 & 0.936 & 0.972 \\
    \bottomrule
    \end{tabular}
\end{table*}

\section{Conclusion}
We propose self-supervised method for to jointly learn depth, optical flow and camera motion.
And we improved the joint self-supervised framework from two aspects: network structure and loss function.
In terms of network architecture, we propose a convolution module for pose estimation based on the attention mechanism. 
Experiments show that our attention module effectively improves the accuracy of pose estimation.
In terms of loss function, benefiting from joint learning we propose a method of dynamic area detection, which effectively eliminates the dynamic area in the loss function.
In addition, we proposed forward and backward optical flow direction constraint and we upgrade the constraint of epipolar geometry.
Finally, compared with other joint self-supervised learning methods, our method achieves SOTA results in optical flow and pose estimation and competitive result in depth estimation.

\ifCLASSOPTIONcaptionsoff
  \newpage
\fi

\bibliographystyle{IEEEtran.bst}
\bibliography{IEEEabrv, ref}

\end{document}